\documentclass[runningheads]{llncs}
\usepackage{graphicx}
\usepackage{algorithmic}
\usepackage{algorithm}
\usepackage{mathrsfs}
\usepackage{amsmath}
\usepackage{amssymb}
\usepackage{hyperref}
\newcommand{\repeatthanks}{\textsuperscript{\thefootnote}}
\begin{document}
\title{Adversarial Policy Gradient for Deep Learning Image Augmentation\thanks{Funded by National Institute of Arthritis and Musculoskeletal and Skin Diseases.}}

\author{Kaiyang Cheng\inst{1, 2}\thanks{These authors contributed equally to this paper.} \and
Claudia Iriondo\inst{1,2}\repeatthanks \and
Francesco Caliv\'a\inst{2} \and
Justin Krogue\inst{3} \and
Sharmila Majumdar\inst{2} \and
Valentina Pedoia\inst{2}}
% index{Cheng, Kaiyang}
% index{Iriondo, Claudia}
% index{Caliva, Francesco}
% index{Krogue, Justin}
% index{Majumdar, Sharmila}
% index{Pedoia, Valentina}

\authorrunning{K. Cheng et al.}

\institute{University of California, Berkeley, USA \\\email{\{victorcheng21, iriondo\}@berkeley.edu} \and
Department of Radiology and Biomedical Imaging, University of California, San Francisco, USA
\email{\{francesco.caliva, sharmila.majumdar, valentina.pedoia\}@ucsf.edu} \and
University of California, San Francisco, USA
\email{justin.krogue@ucsf.edu}
}

\maketitle              % typeset the header of the contribution
\begin{abstract}
The use of semantic segmentation for masking and cropping input images has proven to be a significant aid in medical imaging classification tasks by decreasing the noise and variance of the training dataset. However, implementing this approach with classical methods is challenging: the cost of obtaining a dense segmentation is high, and the precise input area that is most crucial to the classification task is difficult to determine a-priori. We propose a novel joint-training deep reinforcement learning framework for image augmentation. A segmentation network, weakly supervised with policy gradient optimization, acts as an agent, and outputs masks as actions given samples as states, with the goal of maximizing reward signals from the classification network. In this way, the segmentation network learns to mask unimportant imaging features. Our method, Adversarial Policy Gradient Augmentation (APGA), shows promising results on Stanford's MURA dataset and on a hip fracture classification task with an increase in global accuracy of up to 7.33\% and improved performance over baseline methods in 9/10 tasks evaluated. We discuss the broad applicability of our joint training strategy to a variety of medical imaging tasks. 

\end{abstract}
\section{Introduction}
Convolutional neural networks (CNNs) have become an essential part of medical image acquisition, reconstruction, and post-processing pipelines, as this technology significantly improves our ability to detect, study, and predict diseases at scale. In computer-vision, CNNs have achieved above-human performance in natural-object classification tasks~\cite{russakovsky_imagenet_2014}.
However, in medical imaging, where datasets are limited in size and labels are often uncertain, there is still a significant need for methods that maximize information gain while preventing overfitting. Our work presents a novel reinforcement-learning (RL) based image augmentation framework for medical image classification.

Training data image augmentation imposes image-level regularization on CNNs in order to combat overfitting. Augmentation can include the addition of noise, image transformations such as zooming or cropping, image occlusion, and attention masking~\cite{wallenberg_attentional_2017}. The application of the first three methods is limited as they often rely on domain-knowledge to define the appropriate characteristics and severity of the augmentation. The last method requires dense segmentation masks for the region of interest (ROI). However, ROIs relevant to a classification task may not be known \textit{a-priori}. For instance, when inspecting hip radiographs for bone fracture (Fx), the determination of Fx or no-Fx heavily depends on the location of abnormal signal within the bone and abnormalities from nearby tissues. Our RL image augmentation framework leverages an adversarial reward to weakly supervise a segmentation network and create these ROIs.

Through trial-and-error, reinforcement learning algorithms discover how to maximize their objective or reward R. The careful design of reward functions has enabled the application of RL to multiple medical tasks including landmark detection, automatic view planning, treatment planning, and MR/CT reconstruction. In this work, we present APGA, a joint, reinforcement learning strategy for training of a segmentation network and a classification network, that improves accuracy in the classification task.

\section{Methodology}
\subsection{Improving Classification with Segmentation for Image Masking}
The framework has two parts: a classification model with parameters \(\theta_c\) and a segmentation model with parameters (policy) \(\theta_p\). 

To mask out the image-level features that are less useful for the classification task, we use the segmentation model to produce the pixel-wise probability \(p_k\) of the pixel being useful. We zero out the pixels of the original image with \(p_k < 0.5\), and use it for training the classification model, updating \(\theta_c\). With the end-goal of improving the classification performance, our method optimizes the segmentation model to evaluate the importance of each image pixel.
\subsection{Policy Gradient Training}
Following the policy gradient context~\cite{williams_simple_1992}, our segmentation model is seen as a policy in which the image batch is treated as a state $s_t$, whereas the pixel-wise classification of useful image features is framed as an action $a_{t}$. In practice, at each $t$-th training step, the classification model receives the masked image as the input and outputs a reward signal \(R_t\). To accomplish this, our objective is to maximize the expected reward $J(\theta_{p})$ and find the optimal segmentation policy.
\begin{equation} \label{j_theta}
    J(\theta_p) = E_{P(a_t; \theta_p)}[R_t]
\end{equation}
In eq.~\ref{j_theta}, $E_{P(a_t; \theta_p)}$ is the expected reward with respect to the probability of taking an action $a_{t}$, when the model has been parameterized with $\theta_{p}$. 
The policy is learned through back-propagation, which requires the definition of the gradient of the expected reward with respect to the model parameters. Following the REINFORCE rule presented in~\cite{williams_simple_1992}, the gradient can be defined as
\begin{equation} \label{eq. REINFORCE}
    \nabla_{\theta_{p}}J(\theta_p) = E_{P(a_t; \theta_p)}[\nabla_{\theta_{p}}\log(P(a_t | s_t; \theta_p)) \cdot R_t]
\end{equation}
The expected reward cannot be estimated and requires approximation. As is common practice in the tractation of policy gradient, we can achieve such approximation using the negative log-likelihood loss, which is differentiable with respect to the model parameters, and can be properly weighted by the reward signal to obtain the segmentation policy loss presented in eq.~\ref{eq. segmentation_policy}
\begin{equation}\label{eq. segmentation_policy}
    \mathscr{L} = \mathscr{L}_{BCE}(P(a_t | s_t; \theta_p), a_t) \cdot R_t 
\end{equation}
where $\mathscr{L}_{BCE}$ is the binary cross-entropy loss 
\begin{equation} \label{eq: cross-entropy-loss}
\begin{split}
 \mathscr{L}_{BCE}(\widehat{y}, y)=-\frac{1}{N}\sum_{i=1}^{N}y_{i}\cdot & \log (\widehat{y}_{i}) + (1 - y_{i})\cdot\log(1 - \widehat{y}_{i})
\end{split}
\end{equation}
which becomes eq.~\ref{eq. negative_likelihoods}.
\begin{equation} \label{eq. negative_likelihoods}
\begin{split} 
   \mathscr{L}_{BCE}(P(a_t | s_t; \theta_p), a_t)= -\frac{1}{N}\sum_{i=1}^{N} \Big[a_t\cdot & \log(P(a_t | s_t; \theta_p)) + \\ 
   & (1-a_t)\cdot \log(1- P(a_t | s_t; \theta_p))\Big]_{i}
\end{split}
\end{equation}
By using pixel-wise binary cross entropy, we can achieve preservation of spatial information of the deviation of \(P(a_t | s_t)\) from \(a_t\). We then update \(\theta_p\) by computing \(\nabla_{\theta_{p}} \mathscr{L}\).
Consequently, the classification model parameters \(\theta_p\) are updated with gradient descent by using the cross entropy loss between the classification of the masked image samples and the original target labels. In our experiments, we perform stochastic gradient update for both \(\theta_p\) and \(\theta_c\) at each batch step.
\begin{figure}[!t]
\centering
\includegraphics[width=\textwidth]{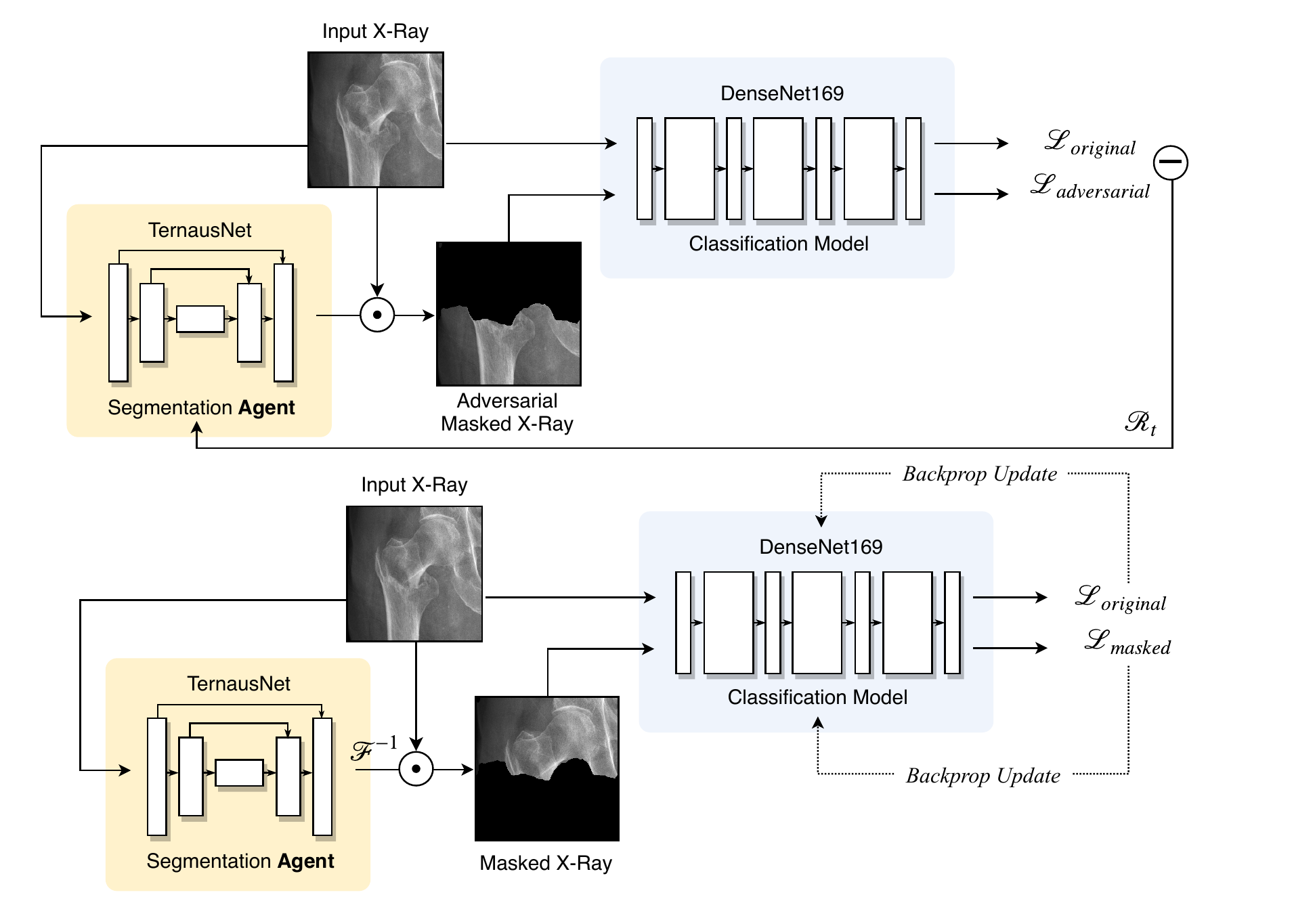}
\caption{ (top) Adversarial training of segmentation agent and (bottom) training of classification model with original and augmented data} \label{fig1}
\end{figure}
\subsection{Adversarial Reward}
 The design of the reward is crucial to the convergence of the segmentation model. Using the change in training loss as a reward, as is done in Neural Architecture Search~\cite{zoph_neural_2016}, results in a weak reward signal hardly discernible from the expected changes in loss during training. Similarly, approximating rewards with a critic network introduces unnecessary overhead and slows down convergence. We propose a stable adversarial reward $R_t$.
 Given pixel-wise feature importance probability $p_k = P(a_t | s_t; \theta_p)$, we zero out the pixels $p_k > 0.5$ to mask-out the features predicted to be of high importance. The original and masked image batches are then fed as inputs to the classification model producing the losses \(\mathscr{L}_{original}\) and \(\mathscr{L}_{adversarial}\). The reward function is defined as:
 \begin{equation}
     R_t = \mathscr{L}_{adversarial} - \mathscr{L}_{original}
 \end{equation}
 To reduce the variance of training, a baseline $b_t$, the exponential moving average of the reward, is included, similarly done in \cite{zoph_neural_2016}.
 Intuitively, by erasing the important features we revert the problem and tend to maximize the gain in loss. However, we do not want the segmentation policy to erase all pixels in favor of a gain in $R_t$, so we penalize the masking of all pixels. Given pixel-wise all zero feature importance  \(a_{zeros}\) and a weight $\lambda_{zeros}$ for regularization, the final
loss \(\mathscr{L}\) is defined as:
\begin{equation}
\begin{split} 
    \mathscr{L} = \mathscr{L}_{BCE}(P(a_t | s_t; \theta_p), a_t)\cdot (R_t - b_t)  + \mathscr{L}_{BCE}(P(a_t | s_t; \theta_p), a_{zeros})\cdot\lambda_{zeros}
\end{split} 
\end{equation}
The resulting reward signal is strongly related to mask quality, rather than reflecting the stochasticity in training of the classification network.  

\begin{algorithm}[t]
  \caption{Adversarial Policy Gradient Augmentation (APGA)}\label{algorithm1}
  %Initialize classification and segmentation models parameters \theta_c and \theta_p;
    \textbf{Input:} Training steps ($T$); training samples and labels ($X$,$Y$); classification and segmentation loss functions ($\mathscr{L}_c$, $\mathscr{L}_p$); learning rates ($\alpha_c$, $\alpha_p$); classification and segmentation models ($M_c$, $M_p$) parameterized with ($\theta_{c}$, $\theta_{p}$).\\
  \textbf{Output:} $M_c$ and $M_p$ 
  \begin{algorithmic}[1]
    \STATE Initialize models parameters \(\theta_c\) and \(\theta_p\)
    \STATE Train classification model on $X$ to convergence
    \FOR{t = 1,..., T}
      \STATE Sample a training batch ($x_t, y_t$) from the ($X, Y$) pool
      \STATE Get classification loss $\mathscr{L}_{original}=\mathscr{L}_c(M_c(x_t), y_t)$
      \STATE Perform gradient descent update $\theta_c \gets \theta_c - \alpha_c \nabla_{\theta_c} \mathscr{L}_{original}$
      \STATE Get adversarial action probabilities from the seg network $P_{adversarial} = M_p(x_t)$
      \STATE Calculate masks of important features $ A_{adversarial} = (P_{adversarial}<0.5) $
      \STATE Produce masks from segmentation networks and erase the predicted features $x_{adversarial} =A_{adversarial} \cdot x_t$
      \STATE Get adversarial loss $\mathscr{L}_{adversarial} = \mathscr{L}_c(M_c(x_{adversarial}), y_t)$
      \STATE Calculate reward $R_t = \mathscr{L}_{adversarial} - \mathscr{L}_{original} $
      \STATE Calculate distances of action probabilities from actions taken $\mathscr{L}_{prob} = \mathscr{L}_p(P_{adversarial}, A_{adversarial})$
      \STATE Calculate distances of action probabilities from the opposite of undesirable extreme actions (masking all pixels) $\mathscr{L}_{extreme} = \mathscr{L}_p(P_{adversarial}, A_{extreme})$
      \STATE Calculate adversarial policy gradient loss $\mathscr{L}_{pg} = \mathscr{L}_{prob} \cdot R_t + \mathscr{L}_{extreme} $
      \STATE Update the segmentation network  $\theta_p \gets \theta_p - \alpha_p \nabla_{\theta_p} \mathscr{L}_{pg}$
      \STATE Produce aiding actions from segmentation network $A_{aid} = M_p(x_t) > 0.5$
      \STATE Update $M_{c}$ with the aiding masks $\theta_c \gets \theta_c - \alpha_c \nabla_{\theta_c} \mathscr{L}_c(M_c(A_{aid}\cdot x_t), y_t)$
    \ENDFOR
  \end{algorithmic}
\end{algorithm}

\section{Experiments and Results}
We evaluate our methodology on MURA~\cite{rajpurkar_mura:_2017} and an internal hip fracture dataset, using the same experimental setup, including network architectures, RL framework, and training hyperparameters. A DenseNet-169~\cite{huang_densely_2016} pretrained on ImageNet~\cite{deng_imagenet:_nodate} serves as the base classification model. A TernausNet~\cite{iglovikov_ternausnet:_2018}, pretrained on Carvana Image Masking Challenge~\cite{noauthor_carvana_nodate}, serves as the segmentation model. Masked images are used as augmentation in a $1:1$ ratio with original images to train the classification network. Images  resized to $224\times224$, batch size 25. Adam optimizers~\cite{kingma_adam:_2014} with a initial learning rate of 0.0001 are used for the classification and the segmentation model. The exponential average baseline $b_t$ has a decay rate of 0.5. $\lambda_{zeros}$ is set to 0.1. Training of APGA converges within 30 minutes on a single Nvidia TitanX GPU. Source code available at \hyperref[]{https://github.com/victorychain/Adversarial-Policy-Gradient-Augmentation}.
\subsubsection{Baselines: }We benchmark APGA using a DenseNet-169~\cite{huang_densely_2016} classifier trained (1) without data augmentation, (2) with cutout~\cite{devries_improved_2017} augmentation in a 1:1 ratio, and with (3) GradCam~\cite{selvaraju_grad-cam:_2016} derived masks augmentation also in a 1:1 ratio.  Cutout augmentation masks out randomly sized patches of the input image while GradCam masks are produced by discretizing the probability saliency map from the DenseNet trained without data augmentation. For further comparison, a segmentation and classification network are trained end-to-end, by propagating the gradient from the classification loss function through the segmentation network and applying the discretized masks from the segmentation network in the same update step. Additionally, regularization terms $a_{zeros}$ and $\lambda_{zeros}$, $a_{ones}$ and $\lambda_{ones}$, are added to the loss function to prevent all or none masking behavior. However, end-to-end training was unstable, and the segmentation network produced all-one or all-zero masks, despite tuning of $\lambda_{zeros}$ and $\lambda_{ones}$. Therefore, these results were omitted. At its best, the end-to-end network produced all-one masks and performed the same as the DenseNet trained without augmentation.
\subsection{Binary Classification: MURA}
The MURA~\cite{rajpurkar_mura:_2017} dataset contains 14,863 musculoskeletal studies of elbows, finger, forearm, hand, humerus, shoulder, and wrist, which contains 9,045 normal and 5,818 abnormal labeled cases. We train the methods on the public training set and evaluate on the validation set, with global accuracy as the metric. We train and evaluate separate models on each body part, and train a single model on a random sample of 100 training images per class to test the performance of our method under extreme data constraints. The performance on the validation set is presented in Table~\ref{mura_table}, Table~\ref{elbow_table} as average and standard deviation of 5 random seeds. 
\begin{table}
\centering
\caption{Classification results (validation accuracy) on MURA.}\label{mura_table}
\begin{tabular}{c|c|c|c|c}
\hline \hline
 &  \ DenseNet & \vtop{\hbox{\strut  DenseNet}\hbox{\strut + cutout}} & \vtop{\hbox{\strut  DenseNet}\hbox{\strut + GradCam}} & \vtop{\hbox{\strut  APGA}\hbox{\strut (Ours)}}\\
\hline
Elbow &  $82.80\pm1.10$ &  $83.05\pm0.85$ & $83.18\pm0.70$ & $\mathbf{83.26\pm0.62}$\\
Finger &  $75.70\pm1.11$ &  $77.66\pm1.69$ & $76.66\pm0.83$ & $\mathbf{77.96\pm0.19}$\\
Forearm &  $80.07\pm1.98$ &  $81.80\pm1.47$ &  $79.80\pm1.70$ & $\mathbf{83.44\pm0.53}$\\
Hand & $78.35\pm0.82$ &  $78.13\pm0.86$ & $77.35\pm1.49$ & $\mathbf{79.09\pm1.44}$\\
Humerus & $85.69\pm0.93$ &  $85.69\pm0.79$ & $86.39\pm1.50$ & $\mathbf{86.53\pm0.67}$\\
Shoulder & $75.88\pm0.93$ &  $75.77\pm0.78$ & $\mathbf{76.80\pm1.18}$ & $76.59\pm0.48$\\
Wrist & $83.28\pm0.46$ &  $84.04\pm0.97$ & $82.61\pm1.19$ & $\mathbf{84.41\pm0.54}$\\
\hline \hline
\end{tabular}
%\end{table}
%\begin{table}
\newline \newline
\centering
\caption{100-shot results (validation accuracy) on Elbow in MURA.}\label{elbow_table}
\begin{tabular}{c|c|c|c|c}
\hline \hline 
 &  \ DenseNet & \vtop{\hbox{\strut  DenseNet}\hbox{\strut + cutout}} & \vtop{\hbox{\strut  DenseNet}\hbox{\strut + GradCam}} & \vtop{\hbox{\strut  APGA}\hbox{\strut (Ours)}}\\
 \hline
Elbow &  $62.68\pm2.29$ &  $64.49\pm2.77$ & $65.87\pm1.97$ & $\mathbf{70.41\pm2.76}$\\
\hline \hline 
\end{tabular}
\end{table}
\begin{figure}
\includegraphics[width=1.97cm]{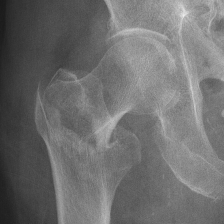}
\includegraphics[width=1.97cm]{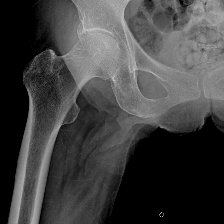}
\includegraphics[width=1.97cm]{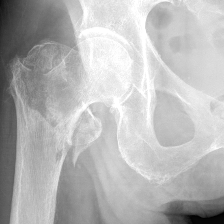}
\includegraphics[width=1.97cm]{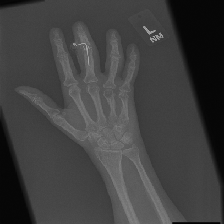}
\includegraphics[width=1.97cm]{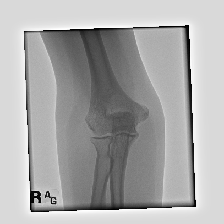}
\includegraphics[width=1.97cm]{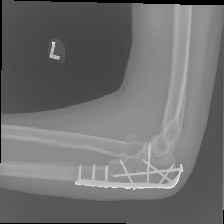}
\newline
\includegraphics[width=1.97cm]{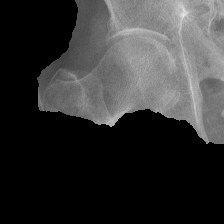}
\includegraphics[width=1.97cm]{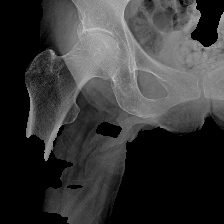}
\includegraphics[width=1.97cm]{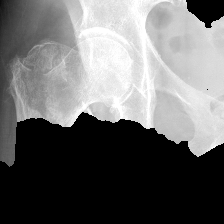}
\includegraphics[width=1.97cm]{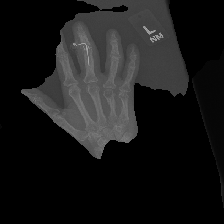}
\includegraphics[width=1.97cm]{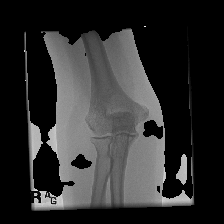}
\includegraphics[width=1.97cm]{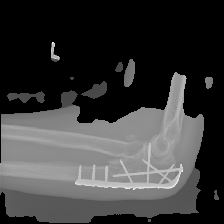}
\newline
\includegraphics[width=1.97cm]{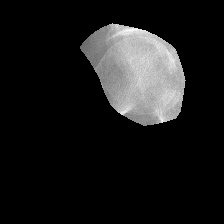}
\includegraphics[width=1.97cm]{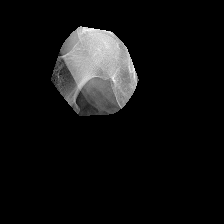}
\includegraphics[width=1.97cm]{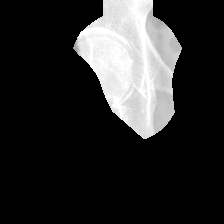}
\includegraphics[width=1.97cm]{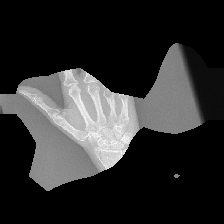}
\includegraphics[width=1.97cm]{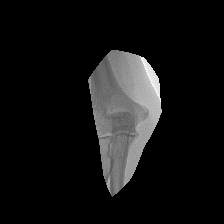}
\includegraphics[width=1.97cm]{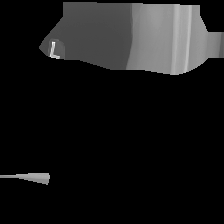}
\caption{ Example X-Rays (top) and masks created by APGA (middle) and DenseNet+GradCam (bottom) for hip, hand, and elbow.} \label{fig2}
\end{figure}
\subsection{Multi-class Classification: Hip Fracture}
The Hip Fracture dataset contains 1118 studies with an average patient age of 74.6 years (standard deviation 17.3), and a
$62:38$ female:male ratio. Each study includes a pelvic radiograph, labeled as 1 of 6 classes: No fracture, Intertrochanteric fracture, Displaced femoral neck fracture, Non-displaced femoral neck fracture, Arthroplasty, or ORIF (previous internal fixation). Bounding boxes are manually drawn on each study, resulting in 3034 bounded hips. The images are split by accession number into train:valid:test  using a $60:25:15$ split, ensuring no overlap in patients between any of the sets. We train and evaluate separate models on the whole pelvic radiographs and the bounded hip radiographs. Per-image accuracy is used as the metric. The performance on the validation and test set is shown in Table~\ref{hip_table}.
\begin{table}
\centering
\caption{Classification results (validation and test accuracy) on Hip Fx Dataset.} \label{hip_table}
\begin{tabular}{c|c|c|c|c}
\hline \hline
 &  \ DenseNet~\cite{huang_densely_2016} & \vtop{\hbox{\strut  DenseNet}\hbox{\strut + cutout}} & \vtop{\hbox{\strut  DenseNet}\hbox{\strut + GradCam}} & \vtop{\hbox{\strut  APGA}\hbox{\strut (Ours)}}\\
\hline
Whole Pelvis (val) &  $59.69\pm1.64$ &  $\mathbf{62.33\pm1.49}$ & $61.09\pm0.52$ & $60.46\pm1.2$\\
Whole Pelvis (test) &  $51.66\pm1.32$ &  $52.51\pm2.58$ & $51.92\pm1.59$ & $\mathbf{53.38\pm2.5}$\\
Bounded Hip (val)&  $86.84\pm0.62$ &  $\mathbf{88.65\pm0.89}$  & $86.51\pm2.28$ & $87.86\pm0.99$\\
Bounded Hip (test)&  $ 84.53\pm1.21$ &  $85.01\pm2.58$  & $83.95\pm2.72$ & $\mathbf{85.65\pm1.53}$\\
\hline \hline
\end{tabular}
\end{table}
\subsubsection{Results: } Compared to the baseline, our method achieved higher global accuracy in 9 out of 10 tasks including binary (MURA Table~\ref{mura_table}) and multi-class (hip Fx Table~\ref{hip_table}) classification tasks. On average, our method improved MURA validation accuracy by 1.56\% and hip validation and testing accuracy by 0.78\% and 1.72\% respectively. The most significant improvement in accuracy over the baseline was 7.33\% and it was achieved in a data-constrained condition, reported in Table~\ref{elbow_table}. In this particular experiment, the elbow training data was limited to 100 samples per class. Overall, APGA outperformed baseline methods in 9 out of 10 tasks, and consistently provided higher testing results. Example segmentation masks from the weakly supervised network are shown in Fig~\ref{fig2}. APGA learns to ignore unimportant features in the radiographs, such as anatomy irrelevant to the classification task. APGA masking appears more exploratory in nature compared to saliency based attention masking (DenseNet + GradCam), which contains biases from the converged model.

\section{Discussions and Conclusions}
We propose a framework, APGA, for producing segmentations to aid medical image classification in a reinforcement learning setting. This framework requires no manual segmentation, which has the benefit of scalability and generalizability. The system is trained online with the goal of improving the performance of the main task, classification. If no improvement is seen, this can be a check for the assumption that masking based augmentation would aid classification, before pursuing more manual work. Marginal improvements should be evidence that APGA has the potential to add valuable information to the training process. The computational overhead in training is justified by those added benefits, and could be eliminated during inference, as the segmentation network can also be used as an inference augmentation technique. This general reinforcement learning with adversarial reward framework could easily be adopted for other medical imaging tasks, involving regression, and segmentation, with different aiding methods, such as bounding box detection, image distortion, and image generation. The reinforcement guided data augmentation has more generalizability compared to traditional data augmentation based on domain knowledge.

\bibliographystyle{splncs04}
\bibliography{paper1773}

\begin{thebibliography}{10}
\providecommand{\url}[1]{\texttt{#1}}
\providecommand{\urlprefix}{URL }
\providecommand{\doi}[1]{https://doi.org/#1}

\bibitem{noauthor_carvana_nodate}
Carvana {Image} {Masking} {Challenge},
  \url{https://kaggle.com/c/carvana-image-masking-challenge}

\bibitem{deng_imagenet:_nodate}
Deng, J., Dong, W., Socher, R., Li, L.J., Li, K., Fei-Fei, L.: {ImageNet}: {A}
  {Large}-{Scale} {Hierarchical} {Image} {Database} p.~8

\bibitem{devries_improved_2017}
DeVries, T., Taylor, G.W.: Improved {Regularization} of {Convolutional}
  {Neural} {Networks} with {Cutout}. arXiv:1708.04552 [cs]  (Aug 2017),
  \url{http://arxiv.org/abs/1708.04552}, arXiv: 1708.04552

\bibitem{huang_densely_2016}
Huang, G., Liu, Z., van~der Maaten, L., Weinberger, K.Q.: Densely {Connected}
  {Convolutional} {Networks}. arXiv:1608.06993 [cs]  (Aug 2016),
  \url{http://arxiv.org/abs/1608.06993}, arXiv: 1608.06993

\bibitem{iglovikov_ternausnet:_2018}
Iglovikov, V., Shvets, A.: {TernausNet}: {U}-{Net} with {VGG}11 {Encoder}
  {Pre}-{Trained} on {ImageNet} for {Image} {Segmentation}. arXiv:1801.05746
  [cs]  (Jan 2018), \url{http://arxiv.org/abs/1801.05746}, arXiv: 1801.05746

\bibitem{kingma_adam:_2014}
Kingma, D.P., Ba, J.: Adam: {A} {Method} for {Stochastic} {Optimization}.
  arXiv:1412.6980 [cs]  (Dec 2014), \url{http://arxiv.org/abs/1412.6980},
  arXiv: 1412.6980

\bibitem{rajpurkar_mura:_2017}
Rajpurkar, P., Irvin, J., Bagul, A., Ding, D., Duan, T., Mehta, H., Yang, B.,
  Zhu, K., Laird, D., Ball, R.L., Langlotz, C., Shpanskaya, K., Lungren, M.P.,
  Ng, A.Y.: {MURA}: {Large} {Dataset} for {Abnormality} {Detection} in
  {Musculoskeletal} {Radiographs}. arXiv:1712.06957 [physics]  (Dec 2017),
  \url{http://arxiv.org/abs/1712.06957}, arXiv: 1712.06957

\bibitem{russakovsky_imagenet_2014}
Russakovsky, O., Deng, J., Su, H., Krause, J., Satheesh, S., Ma, S., Huang, Z.,
  Karpathy, A., Khosla, A., Bernstein, M., Berg, A.C., Fei-Fei, L.: {ImageNet}
  {Large} {Scale} {Visual} {Recognition} {Challenge}. arXiv:1409.0575 [cs]
  (Sep 2014), \url{http://arxiv.org/abs/1409.0575}, arXiv: 1409.0575

\bibitem{selvaraju_grad-cam:_2016}
Selvaraju, R.R., Cogswell, M., Das, A., Vedantam, R., Parikh, D., Batra, D.:
  Grad-{CAM}: {Visual} {Explanations} from {Deep} {Networks} via
  {Gradient}-based {Localization}. arXiv:1610.02391 [cs]  (Oct 2016),
  \url{http://arxiv.org/abs/1610.02391}, arXiv: 1610.02391

\bibitem{wallenberg_attentional_2017}
Wallenberg, M., Forssén, P.: Attentional masking for pre-trained deep
  networks. In: 2017 {IEEE}/{RSJ} {International} {Conference} on {Intelligent}
  {Robots} and {Systems} ({IROS}). pp. 6149--6154 (Sep 2017).
  \doi{10.1109/IROS.2017.8206516}

\bibitem{williams_simple_1992}
Williams, R.J.: Simple statistical gradient-following algorithms for
  connectionist reinforcement learning. Machine Learning  \textbf{8}(3),
  229--256 (May 1992). \doi{10.1007/BF00992696},
  \url{https://doi.org/10.1007/BF00992696}

\bibitem{zoph_neural_2016}
Zoph, B., Le, Q.V.: Neural {Architecture} {Search} with {Reinforcement}
  {Learning}. arXiv:1611.01578 [cs]  (Nov 2016),
  \url{http://arxiv.org/abs/1611.01578}, arXiv: 1611.01578

\end{thebibliography}

\end{document}